\documentclass[conference]{IEEEtran}
\IEEEoverridecommandlockouts

\usepackage{cite}
\usepackage[numbers,sort&compress]{natbib}
\usepackage{amsmath,amssymb,amsfonts}
\usepackage{algorithmic}
\usepackage{graphicx}
\usepackage{textcomp}
\usepackage{xcolor}
\usepackage[percent]{overpic}
\usepackage{booktabs}
\usepackage{threeparttable}
\def\BibTeX{{\rm B\kern-.05em{\sc i\kern-.025em b}\kern-.08em
    T\kern-.1667em\lower.7ex\hbox{E}\kern-.125emX}}
\begin{document}

\title{Multi-Modal Sensor Fusion using Hybrid Attention for Autonomous Driving\\
\thanks{This work is a result of the joint research project STADT:up (19A22006O). The project is supported by the German Federal Ministry for Economic Affairs and Climate Action (BMWK), based on a decision of the German Bundestag. The author is solely responsible for the content of this publication.}
}

\author{\IEEEauthorblockN{Mayank Mayank\IEEEauthorrefmark{1}\IEEEauthorrefmark{2}, Bharanidhar Duraisamy\IEEEauthorrefmark{1}, Florian Geiss\IEEEauthorrefmark{1},  Abhinav Valada\IEEEauthorrefmark{2} \\
\IEEEauthorblockA{\IEEEauthorrefmark{1} Research and Development, Mercedes-Benz AG, Germany, Email: [firstname].[lastname]@mercedes-benz.com}
\IEEEauthorblockA{\IEEEauthorrefmark{2} Department of Computer Science, University of Freiburg, Germany, Email : valada@cs.uni-freiburg.de}
}}

\maketitle

\begin{abstract}
Accurate 3D object detection for autonomous driving requires complementary sensors. Cameras provide dense semantics but unreliable depth, while millimeter-wave radar offers precise range and velocity measurements with sparse geometry. We propose \textbf{MMF-BEV}, a radar–camera BEV fusion framework that leverages deformable attention for cross-modal feature alignment on the View-of-Delft (VoD) 4D radar dataset~\cite{palffy2022multi}. MMF-BEV builds a BEVDepth ~\cite{li2023bevdepth} camera branch and a RadarBEVNet~\cite{lin2024rcbevdet} radar branch, each enhanced with Deformable Self-Attention, and fuses them via a Deformable Cross-Attention module. We evaluate three configurations: camera-only, radar-only, and hybrid fusion. A sensor contribution analysis quantifies per-distance modality weighting, providing interpretable evidence of sensor complementarity. A two-stage training strategy - pre-training the camera branch with depth supervision, then jointly training radar and fusion modules stabilizes learning. Experiments on VoD show that MMF-BEV consistently outperforms unimodal baselines and achieves competitive results against prior fusion methods across all object classes in both the full annotated area and near-range Region of Interest.
\end{abstract}


\section{Introduction}
\label{sec:intro}

Three-dimensional object detection is a safety-critical task in autonomous driving. Among available sensor modalities, cameras offer rich texture and semantic detail but fundamentally cannot measure depth directly~\cite{ma2021delving}, making them sensitive to depth estimation errors in BEV-based detectors. Millimeter-wave radar sensors complement this weakness with precise range measurements via time-of-flight and radial velocity via the Doppler effect; they also operate reliably in adverse lighting and weather conditions~\cite{zhou2022towards}. However, radar point clouds are geometrically sparse and carry no semantic information, making radar-only detection insufficient for reliable perception. A fusion approach that exploits both modalities in a shared BEV representation is therefore both practically appealing and technically well-motivated.

The Lift-Splat-Shoot (LSS) framework~\cite{philion2020lift} established a canonical pipeline for projecting image features into Bird’s Eye View (BEV) using predicted depth distributions. Building on this, BEVDepth~\cite{li2023bevdepth} improved depth estimation by incorporating explicit LiDAR supervision and a camera-aware depth prediction module, providing a strong camera-only baseline. RCBEVDet~\cite{lin2024rcbevdet} extended BEV fusion to radar by introducing RadarBEVNet, a dual-stream radar encoder with Radar Cross-Section (RCS)-aware BEV scattering, and a Cross-Attention Multi-layer Fusion (CAMF) module that leverages deformable cross-attention~\cite{zhu2020deformable} to align camera and radar features in BEV space.

Despite the progress in BEV-based camera and radar fusion, two questions remain under-explored on the challenging View-of-Delft (VoD) 4D radar dataset~\cite{palffy2022multi}: (i) the quantitative impact of intra-modal deformable self-attention (DSA) on each modality independently, and (ii) how cross-modal attention attends to camera and radar features across object classes and distances. VoD’s emphasis on small, vulnerable road users (pedestrians and cyclists) and its use of elevation-resolved 4D radar measurements make it a particularly suitable benchmark for studying these effects. To address these limitations, we propose \textbf{MMF-BEV}, a radar–camera BEV fusion framework that systematically evaluates deformable self- and cross-attention mechanisms while providing interpretable sensor contribution analysis. Its main contributions are summarized as follows:

\begin{itemize}

    \item \textbf{Systematic study of deformable self-attention (DSA).}
    We investigate DSA as an intra-modal refinement mechanism for both sensing branches, augmenting the camera BEV encoder (BEVDepth + DSA) and the radar BEV encoder (RadarBEVNet + DSA). Each enhanced branch is evaluated independently as a standalone 3D detector on the VoD benchmark to quantify the isolated effect of modality-specific deformable attention.

    \item \textbf{MultiLayer Hybrid Fusion.}
    We propose a symmetric cross-modal fusion module in which (i) camera BEV features query radar BEV features and (ii) radar BEV features query camera BEV features. Cross-attention is applied after per-modality DSA refinement and followed by channel- and spatial-aware CBR-based feature aggregation, enabling structured and spatially adaptive inter-modal interaction.

    \item \textbf{Sensor contribution analysis in BEV space.}
    We introduce an interpretable analysis framework that normalizes camera and radar BEV feature magnitudes at each spatial bin and estimates their relative contribution per detection. The resulting modality weights are analyzed across object classes (Car, Pedestrian, Cyclist) and radial distance intervals on the VoD validation set, providing insight into how the model allocates reliance between modalities.

    \item \textbf{Two-stage training strategy for stable fusion.}
    We adopt a staged optimization procedure: first training the BEVDepth-based camera branch with explicit depth supervision on VoD, followed by training the radar branch and fusion module while freezing camera parameters. Comprehensive results are reported for both the entire annotated area and the near-range Region of Interest (ROI).

\end{itemize}

\section{Related Work}
\label{sec:related}

\subsection{Camera-Based BEV Detection}

\begin{figure}[t]
  \centering
  \includegraphics[
    width=1.0\linewidth,
    trim=90 500 90 50,  
    clip
  ]{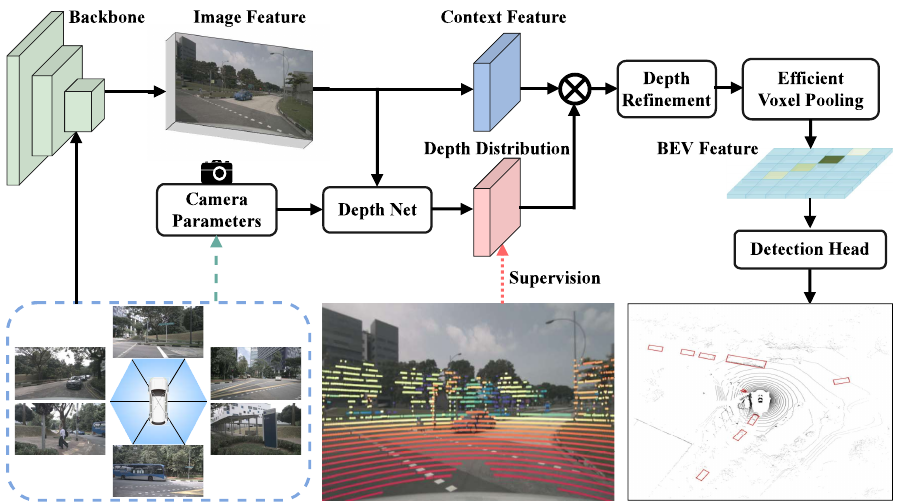}
  \caption{BEVDepth Framework \cite{li2023bevdepth}}
  \label{fig:bevdepth_framework}
\end{figure}

The Lift-Splat-Shoot (LSS) paradigm~\cite{philion2020lift} transforms per-pixel depth distributions into 3D frustum features pooled into a BEV map. BEVDet~\cite{huang2021bevdet} applied this for 3D detection with BEV-space augmentations; BEVDet4D~\cite{huang2022bevdet4d} added temporal alignment of consecutive BEV frames. BEVDepth~\cite{li2023bevdepth} showed that implicit lidar depth supervision as in Fig.~\ref{fig:bevdepth_framework} from detection loss produces surprisingly poor depth estimates—even a detector achieving 30 mAP can have an AbsRel depth error of 3.03—and resolved this with explicit LiDAR-derived depth supervision, a camera-intrinsic-aware DepthNet, and a Depth Refinement Module, boosting mAP by nearly 20 points when using ground-truth depth. BEVDepth also introduced Efficient Voxel Pooling, an 80$\times$ faster GPU-parallel alternative to the cumulative-sum trick. BEVFormer~\cite{li2022bevformer} instead used spatiotemporal BEV queries with deformable attention~\cite{zhu2020deformable} to aggregate image features without explicit depth prediction. PETR~\cite{liu2022petr} and PETRv2~\cite{liu2023petrv2} embedded 3D positional encodings directly into image features, while StreamPETR~\cite{wang2023exploring} propagated object queries across frames for efficient temporal modeling.

\subsection{Radar-Camera Fusion}

\begin{figure}[t]
  \centering
  \includegraphics[
    width=1.0\linewidth,
    trim=50 568 300 50,  
    clip
  ]{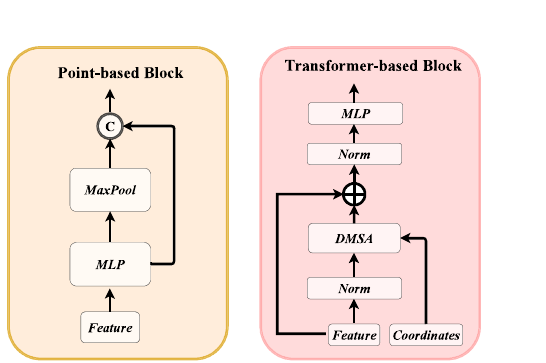}
  \caption{Architecture of the RCBEVDET's dual-stream radar backbone (Pointbased
Block and Transformer-based Block) \cite{lin2024rcbevdet}.}
  \label{fig:rcbevdet_dualstreamradar}
\end{figure}

Radar sensors are cheap, weather-robust, and provide Doppler velocity, motivating their use as a complementary modality. CenterFusion~\cite{nabati2021centerfusion} associated radar points to camera detections via frustum-based matching for refinement. CRAFT~\cite{kim2023craft} introduced a Soft-Polar-Association transformer for proposal-level radar-camera fusion. CRN~\cite{kim2023crn} used radar occupancy to augment camera view transformation and incorporated cross-attention for BEV alignment. RCFusion~\cite{zheng2023rcfusion} generated radar pseudo-images via a radar PillarNet and fused with camera BEV features. RCBEVDet~\cite{lin2024rcbevdet}, the primary baseline for this work, introduced a purpose-built RadarBEVNet with a dual-stream (point-based + transformer-based) backbone as shown in Fig.~\ref{fig:rcbevdet_dualstreamradar}, RCS-aware BEV scattering, and a deformable cross-attention CAMF module. On VoD, RCBEVDet achieved 69.8\% mAP in the ROI, outperforming prior fusion methods. Our MMF-BEV builds on this foundation, adding per-modality DSA refinement along with Deformable Cross Attention, and provides an interpretable sensor contribution analysis absent in prior work.

\subsection{Deformable Attention}

Standard self-attention exhibits quadratic complexity with respect to
the spatial resolution, leading to $\mathcal{O}(H^{2}W^{2}C)$ operations
for BEV features of height $H$, width $W$, and channel dimension $C$.
To mitigate this computational burden, Deformable DETR~\cite{zhu2020deformable}
introduces deformable attention, which restricts aggregation to a sparse
set of $K$ learnable reference sampling points instead of attending to
all spatial locations. As a result, the complexity is reduced to
$\mathcal{O}(HWC^{2}K)$.

\begin{figure*}[t]
  \centering
  \includegraphics[width=\textwidth]{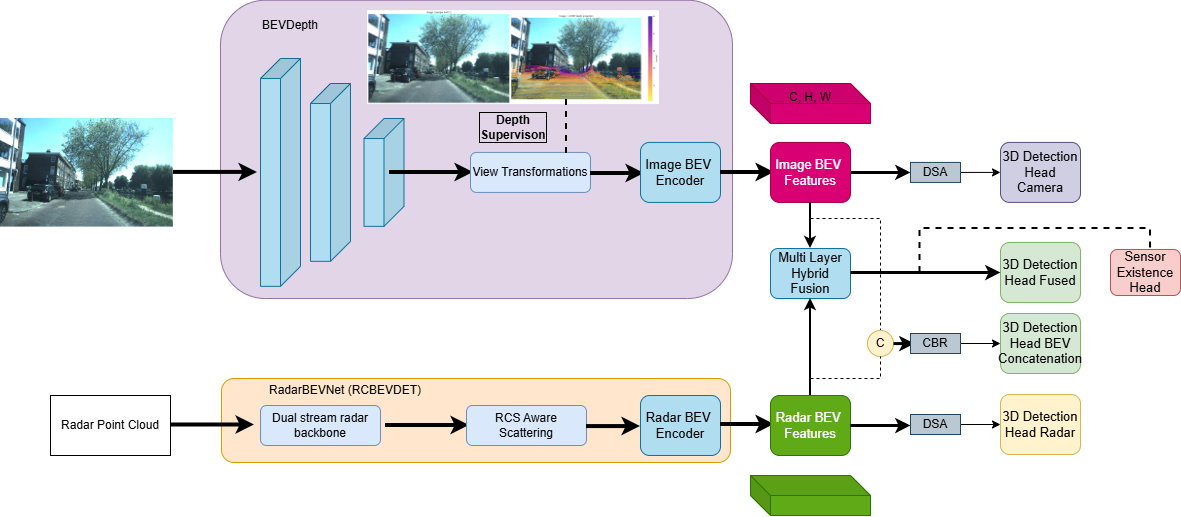}
  \caption{Overall pipeline of MMF-BEV. Front-view camera images are transformed into BEV features using a BEVDepth-based camera branch, while 4D radar point clouds are encoded via RadarBEVNet to produce radar BEV representations. Both modalities are refined with Deformable Self-Attention and fused through  Multi Layer hybrid fusion module in BEV space. The fused multi-modal BEV features are used for 3D object detection and sensor specific confidence for detections.}
  \label{fig:arch}
\end{figure*}

Beyond efficiency, the learnable spatial offsets provide an important
geometric advantage: they enable dynamic compensation for spatial
misalignment. This property is particularly beneficial in MMF-BEV,
where deformable self-attention (DSA) operates within each modality and
deformable cross-attention performs inter-modal fusion. In this setting,
radar azimuth inaccuracies and monocular camera depth ambiguities induce
distinct yet complementary spatial distortions, which the deformable
attention mechanism can adaptively correct through learned offsets.

\subsection{View-of-Delft Dataset}

VoD~\cite{palffy2022multi} provides synchronized measurements from a stereo
camera pair and a Continental ARS548 4D imaging radar (range, azimuth,
elevation, and Doppler) recorded in urban traffic scenarios in Delft.
The dataset includes annotations for cars, pedestrians, and cyclists,
with IoU thresholds of 0.5 for cars and 0.25 for pedestrians and cyclists.
Evaluation is conducted over both the full annotated area and a
driving-corridor region of interest (ROI) defined as 20\,m in front of
the ego-vehicle and $[-4,\,4]$\,m laterally.

The inclusion of the radar elevation dimension and the dataset’s
particular focus on vulnerable road users in close-range urban
environments distinguish VoD from the more widely used
nuScenes~\cite{caesar2020nuscenes} benchmark.

\section{MMF-BEV: Implementation Details}
\label{sec:method}

Fig.~\ref{fig:arch} illustrates the overall MMF-BEV pipeline. The camera branch transforms a front-view image into a camera BEV feature map $F_c$ via BEVDepth \cite{li2023bevdepth}, the radar branch encodes 4D radar point clouds into a radar BEV feature map $F_r$ via RadarBEVNet \cite{lin2024rcbevdet}, and the multi layer hybrid fusion module merges $F_c$ and $F_r$ through deformable attention before a shared CenterPoint~\cite{yin2021center} detection head. Each branch is additionally enhanced with a Deformable Self-Attention (DSA) block applied to its BEV encoder output before concatenation.

\subsection{Camera Branch: BEVDepth with DSA}

Given a front-view image $I$, a ResNet-50~\cite{he2016deep} backbone extracts 2D features $F^{2d}$. The camera-aware DepthNet predicts a depth distribution:
\begin{equation}
  D^{pred} = \psi\!\left(SE\!\left(F^{2d} \,\big|\, \mathrm{MLP}(\xi(R)\oplus\xi(t)\oplus\xi(K))\right)\right),
\end{equation}
conditioned on intrinsics $K$ and extrinsics $(R,t)$ through Squeeze-and-excitation blocks \cite{li2023bevdepth}. Explicit depth supervision is applied with binary cross-entropy against LiDAR-projected ground-truth depth:
\begin{equation}
  \mathcal{L}_{depth} = \mathrm{BCE}(D^{pred}, D^{gt}).
\end{equation}
The features of the points $F^{3d} = F^{2d} \otimes D^{pred}$ are not projected onto the ego-space and are pooled into $F_c^{bev}$ via efficient vector pooling, followed by a Depth Refinement Module. A Deformable Self-Attention block is then applied to $F_c^{bev}$:
\begin{equation}
  \hat{F}_c = \mathrm{DeformSelfAttn}(F_c^{bev}),
\end{equation}
refining spatial features by attending to learned sparse reference offsets within the camera BEV space.

\textbf{VoD configuration.} A single forward-facing camera is used with an
input resolution of $512 \times 800$. The BEV spatial extent is defined as
$[-51.2,\, 51.2]$\,m in the lateral direction and $[0,\, 51.2]$\,m in the
forward direction, discretized into a $128 \times 128$ BEV grid.

\subsection{Radar Branch: RadarBEVNet with DSA}

Aggregated raw 4D radar point measurements from five consecutive sweeps are
represented as $(x, y, z, v_{x}^{\mathrm{comp}}, v_{y}^{\mathrm{comp}},
\mathrm{RCS}, t)$ and processed using RadarBEVNet~\cite{lin2024rcbevdet}.
Here, $t$ denotes the timestamp associated with each radar point during
temporal accumulation.

\textbf{Dual-stream backbone.} A point-based stream (PointNet-style MLP + global max-pool, $S=3$ blocks) captures local geometric structure, while a transformer-based stream with Distance-Modulated Self-Attention (DMSA) captures global context:
\begin{equation}
  \mathrm{DMSA}(Q,K,V) = \mathrm{Softmax}\!\left(\tfrac{QK^\top}{\sqrt{d}} - \beta D^2\right)\!V,
\end{equation}
where $D$ is the pairwise point-distance matrix and $\beta$ is learnable. An Injection and Extraction module uses cross-attention to couple the two streams at each block.

\textbf{RCS-aware BEV encoder.} Each radar point scatters its feature over a disc of radius $r = \sqrt{c_x^2+c_y^2}\cdot v_{RCS}$ in BEV, with a Gaussian weight map $G_{x,y}$, giving a dense radar BEV feature $f_{RCS}'$. A SECOND-style encoder ~\cite{yan2018second} produces the final radar BEV map $F_r^{bev}$. 

\begin{figure}[t]
  \centering
  \includegraphics[
    width=1.0\linewidth,
    trim=300 285 420 150,  
    clip
  ]{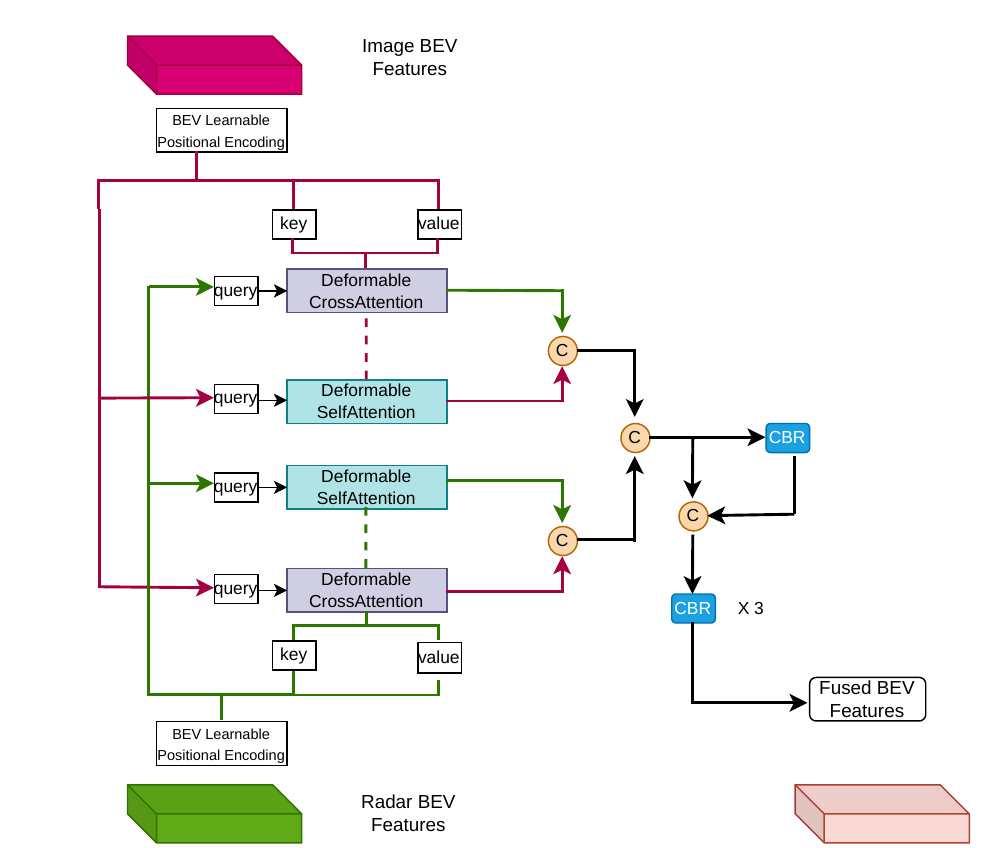}
  \caption{MultiLayer Hybrid fusion module. After per-modality DSA refinement, camera features and radar features exchange information through  deformable cross-attention (top and bottom), with outputs concatenated through CBR fusion layers.}
  \label{fig:attn_head}
\end{figure}

\subsection{MultiLayer Hybrid Fusion Module}

The proposed hybrid fusion module is illustrated in Fig.~\ref{fig:attn_head}.
Let $\hat{F}_c$ and $\hat{F}_r$ denote the camera and radar BEV features,
respectively, after per-modality refinement using deformable self-attention
(DSA) and the addition of learnable BEV positional encodings.
To enable bidirectional information exchange between modalities, we apply
deformable cross-attention (DCA) similar to Deformable DETR\cite{zhu2020deformable} as

\begin{align}
  F_c' &= \mathrm{DeformCrossAttn}(z_{q\hat{c}},\, p_{q\hat{c}},\, \hat{F}_r), \label{eq:ca_c}\\
  F_r' &= \mathrm{DeformCrossAttn}(z_{q\hat{r}},\, p_{q\hat{r}},\, \hat{F}_c), \label{eq:ca_r}
\end{align}

where Eq.~\eqref{eq:ca_c} allows camera queries $z_{q\hat{c}}$ with corresponding 
reference points $p_{q\hat{c}}$ to attend to radar keys and values 
$\hat{F}_r$, thereby grounding semantically rich but depth-ambiguous image 
features to spatially precise radar representations. Conversely, 
Eq.~\eqref{eq:ca_r} enables radar queries $z_{q\hat{r}}$ and reference 
points $p_{q\hat{r}}$ to sample camera features $\hat{F}_c$, enriching 
geometrically accurate yet semantically sparse radar responses with dense 
visual context.

For $M$ attention heads and $K$ learnable reference points per head, the
deformable cross-attention operator is defined as

\begin{equation}
\mathrm{DCA}(z_q, p_q, F) =
\sum_{m=1}^{M} W_m
\sum_{k=1}^{K}
A_{mqk} \cdot
W_m' F\!\left(p_q + \Delta p_{mqk}\right),
\end{equation}

where $\Delta p_{mqk}$ are learnable spatial offsets predicted from the
query features, and $A_{mqk}$ are normalized attention weights. The learned
offsets enable adaptive spatial sampling, which is particularly important
for compensating geometric misalignment between radar and camera modalities.

\paragraph{Joint Self- and Cross-Attention Fusion}
While deformable cross-attention enables complementary information exchange between modalities, it does not explicitly guarantee the preservation of modality-specific structures, particularly in regions where one sensing modality provides weak or uncertain evidence (e.g., radar sparsity or monocular depth ambiguity). To address this limitation, we explicitly retain the modality-specific contextual features obtained from Deformable Self-Attention (DSA) and fuse them jointly with the cross-attended representations.

Let $F_c$ and $F_r$ denote the camera and radar BEV features before attention.  
We define deformable self-attention (with superscript $s$) as
\begin{equation}
F_{c}^{s} = \mathrm{DeformSelfAttn}(F_c),
\end{equation}

\begin{equation}
F_{r}^{s} = \mathrm{DeformSelfAttn}(F_r),
\end{equation}
where $F_{c}^{s}$ and $F_{r}^{s}$ encode modality-specific contextual reasoning. To preserve both modality-specific and cross-modal information, we concatenate
the self-attended and cross-attended features:
\begin{equation}
F_{\mathrm{concat}} =
\mathrm{Concat}(F_{c}^{s}, F_{r}^{s}, F_{c}', F_{r}').
\end{equation}

The final fused BEV representation is obtained through three consecutive
CBR (Conv–BatchNorm–ReLU) blocks \cite{lin2024rcbevdet}:
\begin{equation}
F_{\mathrm{fused}} =
\mathrm{CBR}^{3}\!\left(F_{\mathrm{concat}}\right).
\end{equation}

This hybrid strategy combines:

\begin{itemize}
\item \textbf{Intra-modal consistency} via deformable self-attention, which
captures long-range spatial dependencies within each modality.
\item \textbf{Inter-modal complementarity} via deformable cross-attention,
which aligns and integrates heterogeneous features.
\end{itemize}

By jointly leveraging both mechanisms, the fusion module reduces the risk
of information loss in unattended regions during cross-attention and
improves robustness to modality-specific noise. In practice, this results
in sharper object localization (guided by radar geometry) while preserving
rich semantic discrimination (from camera features), yielding a more
spatially coherent and semantically expressive fused BEV representation.

\subsection{Sensor Contribution Analysis}
\label{subsec:contrib}
We compute per-bin sensor contribution maps over the VoD validation set to interpret how each modality influences final detection outcomes. For each BEV grid cell $(i,j)$, we extract the L2-norm of the camera and radar BEV features:
\begin{equation}
  \bar{F}_c(i,j) = \left\|\hat{F}_c(i,j)\right\|_2, \qquad \bar{F}_r(i,j) = \left\|\hat{F}_r(i,j)\right\|_2.
\end{equation}
The relative camera and radar weights are then:
\begin{equation}
  C(i,j) = \frac{\bar{F}_c(i,j)}{\bar{F}_c(i,j) + \bar{F}_r(i,j)}, \quad R(i,j) = 1 - C(i,j).
\end{equation}
We bin detections by object class and by radial distance from the ego vehicle and compute the mean $C$ and $R$ per bin. This yields class-stratified and distance-stratified sensor contribution maps, providing interpretable evidence of complementary sensor roles. For example, we expect camera features to dominate at short range (where depth estimation is more reliable and objects subtend large image regions) and for pedestrians (where camera texture is discriminative), while radar dominates at longer ranges and for vehicles (which produce strong RCS returns). We considered this head as one of the vital feature that enables test and validation of required criteria to fulfil various rating guidelines outlined in ISO/PAS 8800:2024~\cite{ISO/PAS_8800:2024}.

\section{Training Strategy}
\label{sec:training}

To stabilize optimization and prevent interference between depth estimation
and cross-modal alignment, we adopt a two-stage training procedure that
decouples camera depth learning from radar–camera fusion as in \cite{lin2024rcbevdet}.

\subsection{Stage 1: Camera Branch Pre-training}

In the first stage, the BEVDepth-based camera branch augmented with
Deformable Self-Attention (DSA) is trained independently on the VoD
training split for 12 epochs. The implementation is built upon the
MMDetection3D framework~\cite{mmdet3d2020}; however, we develop a custom
dataset interface tailored to the View-of-Delft (VoD) benchmark,
including adapted data loading, preprocessing, and evaluation routines
to ensure compatibility with the 4D radar format and the official VoD
metrics. The objective function is defined as
\begin{equation}
\mathcal{L}_{\mathrm{cam}} =
\mathcal{L}_{\mathrm{depth}} +
\mathcal{L}_{\mathrm{cls}} +
\mathcal{L}_{\mathrm{box}} +
\mathcal{L}_{\mathrm{vel}},
\end{equation}

where $\mathcal{L}_{\mathrm{depth}}$ supervises depth estimation,
$\mathcal{L}_{\mathrm{cls}}$ is the CenterPoint heatmap classification loss,
$\mathcal{L}_{\mathrm{box}}$ denotes bounding-box regression loss, and
$\mathcal{L}_{\mathrm{vel}}$ corresponds to velocity regression.

Optimization is performed using AdamW~\cite{loshchilov2017decoupled}
with an initial learning rate of $2 \times 10^{-4}$ and cosine annealing
schedule. Image-space augmentations include random horizontal flipping and
random crop–resize operations. In the BEV domain, random flipping and
rotation are applied to enhance geometric invariance.

\subsection{Stage 2: Radar–Camera Fusion Training}

In the second stage, the pre-trained camera branch weights are loaded and
kept frozen to preserve learned depth representations. The radar branch
(RadarBEVNet) and the proposed multilayer hybrid fusion module are then
jointly optimized for 12 epochs.

The fusion training objective is defined as
\begin{equation}
\mathcal{L}_{\mathrm{fuse}} =
\mathcal{L}_{\mathrm{cls}} +
\mathcal{L}_{\mathrm{box}} +
\mathcal{L}_{\mathrm{vel}} 
\end{equation}

We employ AdamW with an initial learning rate of
$1 \times 10^{-4}$ and cosine annealing.
Five temporally accumulated radar sweeps are used as input to improve motion
awareness and spatial density. To enhance robustness against radar sparsity
and measurement noise, radar-specific augmentations are applied, including
random point dropout and additive noise perturbation.

This staged optimization strategy ensures stable convergence by first
learning reliable monocular depth representations and subsequently focusing
on cross-modal alignment and fusion without changing the weights of camera branch.


\section{Experiments}
\label{sec:experiments}

\subsection{Dataset and Metrics}

The View-of-Delft (VoD) dataset~\cite{palffy2022multi} comprises 8,693 annotated frames, partitioned into 5,139 training, 1,296 validation, and 2,258 test samples. Each frame provides synchronized measurements from a forward-facing monocular camera and a 4D millimeter-wave radar, the latter delivering range, azimuth, elevation, and Doppler velocity information.

We adhere to the official evaluation protocol and report per-class Average Precision (AP) as well as mean Average Precision (mAP) for three object categories— Cars (IoU $\geq 0.5$), Pedestrians, and Cyclists (IoU $\geq 0.25$). Our analysis presents results for two distinct regions \cite{palffy2022multi}:
\begin{enumerate}
    \item \textbf{Entire Annotated Region:} This encompasses the full camera Field of View (FoV) up to 50 meters.
    \item \textbf{Driving Corridor:} A more safety-relevant region, defined as a rectangular area on the ground plane directly in front of the ego-vehicle. Its boundaries in camera coordinates are specified as:
    \[
    [-4 \, \text{m} < x < +4 \, \text{m}, \quad z < 25 \, \text{m}]
    \]
\end{enumerate}

\subsection{Main Detection Results}
\label{subsec:main_results}

Tables~\ref{tab:vod_full} and~\ref{tab:vod_roi} compare three MMF-BEV
configurations against published baselines on the VoD validation set:
\textbf{C-only} (BEVDepth + DSA, Stage~1 only),
\textbf{R-only} (RadarBEVNet + DSA, trained end-to-end), and
\textbf{C+R} (full MMF-BEV fusion, Stage~2).
\begin{table}[htbp]
\centering
\caption{Results on Validation Set — \textbf{Entire Annotated Area} (\%).}
\label{tab:vod_full}
\renewcommand{\arraystretch}{1.15}
\begin{threeparttable}
\begin{tabular}{lcccccc}
\toprule
\textbf{Method} & \textbf{In} & \textbf{Car} & \textbf{Ped.} & \textbf{Cyc.} & \textbf{mAP} \\
\midrule
PointPillars~\cite{lang2019pointpillars} & R & 37.06 & 35.04 & 63.44 & 45.18 \\
RCFusion~\cite{zheng2023rcfusion}        & C+R & \textbf{41.70} & \textbf{38.95} & 68.31 & 49.65 \\
RCBEVDet~\cite{lin2024rcbevdet}          & C+R & 40.63 & 38.86 & \textbf{70.48} & \textbf{49.99} \\
\midrule
MMF-BEV  & C   & 16.28 & 15.14 & 35.37 & 22.26 \\
MMF-BEV  & R   & 35.55 & 33.05 &  61.20 & 43.26 \\
MMF-BEV* & C+R & 39.65 & 32.00 & 64.71 & 45.45 \\
MMF-BEV  & C+R & 40.10 & 37.80 & 68.88 & 48.92 \\
\bottomrule
\end{tabular}
\begin{tablenotes}
    \footnotesize
    \item[$^*$] Directly concatenating image and radar BEV features.
\end{tablenotes}
\end{threeparttable}
\end{table}

\begin{table}[htbp]
\centering
\caption{Results on Validation Set — \textbf{Driving Corridor / ROI} (\%).}
\label{tab:vod_roi}
\renewcommand{\arraystretch}{1.15}
\begin{threeparttable}
\begin{tabular}{lcccccc}
\toprule
\textbf{Method} & \textbf{In} & \textbf{Car} & \textbf{Ped.} & \textbf{Cyc.} & \textbf{mAP} \\
\midrule
PointPillars~\cite{lang2019pointpillars} & R & 70.15 & 47.22 & 85.07 & 67.48 \\
RCFusion~\cite{zheng2023rcfusion}        & C+R & 71.87 & 47.50 & \textbf{88.33} & 69.23 \\
RCBEVDet~\cite{lin2024rcbevdet}          & C+R & \textbf{72.48} & \textbf{49.89} & 87.01 & \textbf{69.80} \\
\midrule
MMF-BEV  & C   & 55.01 & 15.78 & 43.67 & 38.15 \\
MMF-BEV  & R   & 67.21 & 44.48 & 78.65 & 63.44 \\
MMF-BEV$^*$  & C+R & 70.36 & 39.23 & 87.27 & 65.62 \\
MMF-BEV  & C+R & 72.21 & 48.23 & 87.20  & 69.21 \\
\bottomrule
\end{tabular}
\begin{tablenotes}
    \footnotesize
    \item[$^*$] Directly concatenating image and radar BEV features.
\end{tablenotes}
\end{threeparttable}
\end{table}

\begin{figure*}[t]
  \centering
  \begin{minipage}[b]{0.26\linewidth}
    \centering
    \includegraphics[width=\linewidth, trim={0mm 0mm 0mm 0mm}, clip]{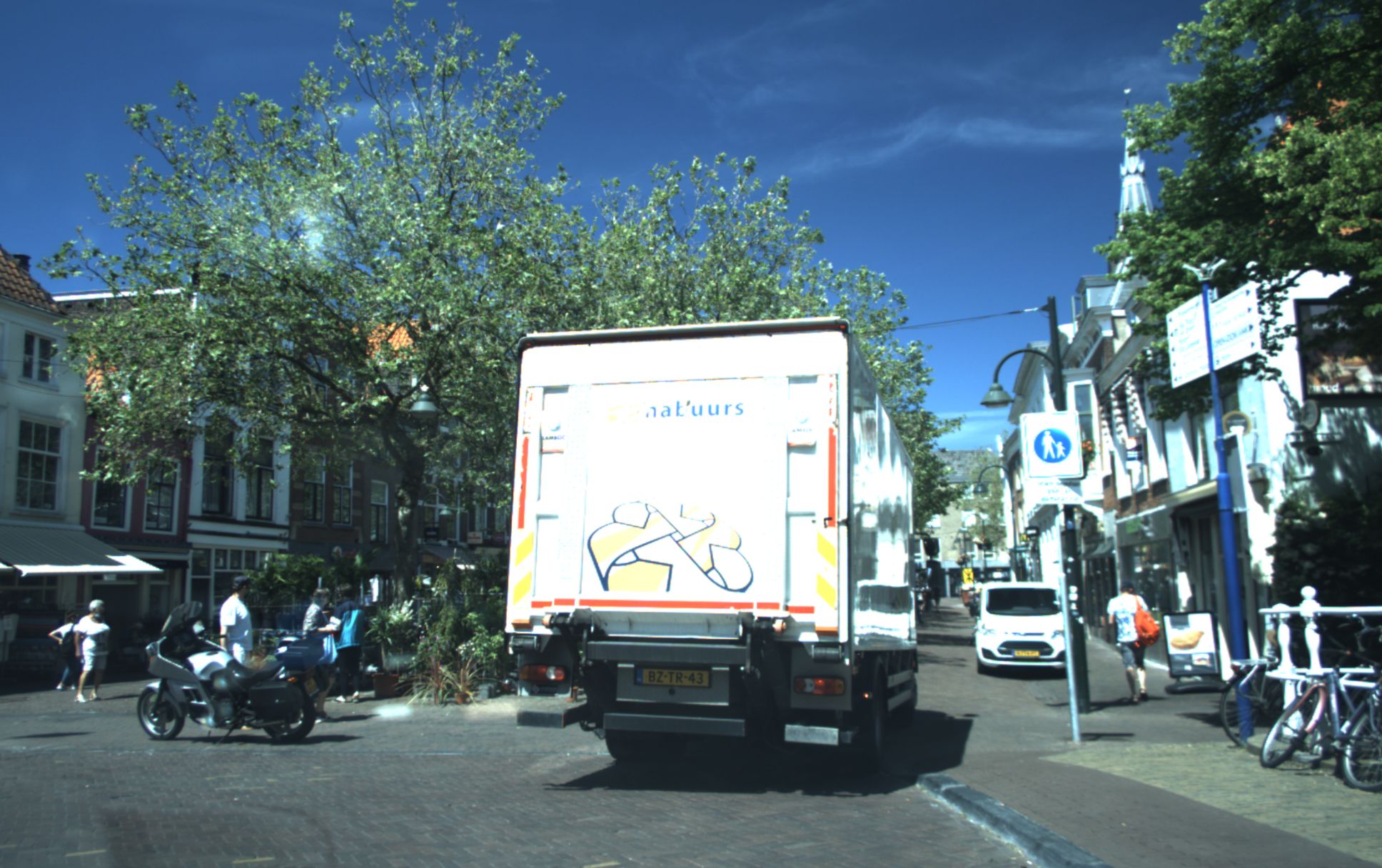}
    \vspace{2pt}
    \small (a) Validation Scene Id - 00000
  \end{minipage}
  \hfill
  \begin{minipage}[b]{0.24\linewidth}
    \centering
    \includegraphics[width=\linewidth, trim={0mm 75mm 35mm 60mm}, clip]{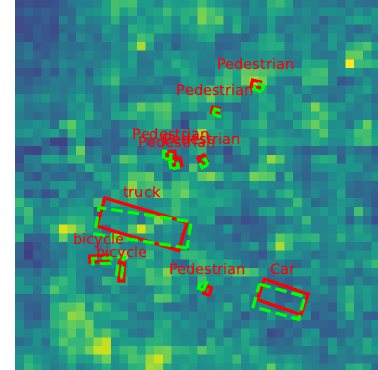}
    \label{fig:bev_qual_b}
    \vspace{2pt}
    \small (b) $F_c'$ - DCA Camera
  \end{minipage}
  \hfill
  \begin{minipage}[b]{0.24\linewidth}
    \centering
    \includegraphics[width=\linewidth, trim={0mm 75mm 35mm 60mm}, clip]{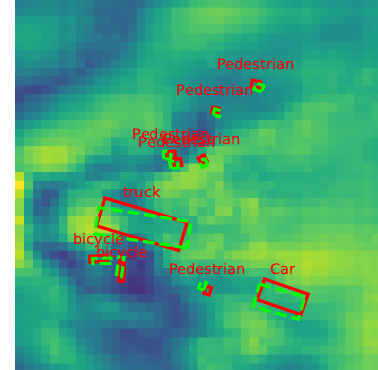}
    \label{fig:bev_qual_c}
    \vspace{2pt}
    \small (c) $F_r'$ - DCA Radar
  \end{minipage}
  \hfill
  \begin{minipage}[b]{0.24\linewidth}
    \centering
    \includegraphics[width=\linewidth, trim={0mm 75mm 35mm 60mm}, clip]{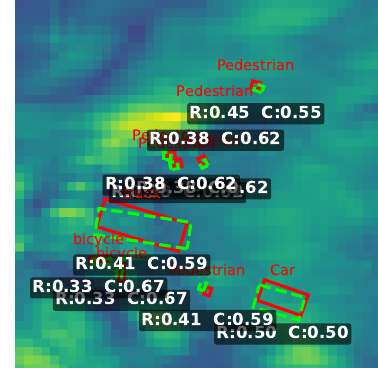}
    \label{fig:bev_qual_d}
    \vspace{2pt}
    \small (d) $F_f$ - Fused Features
  \end{minipage}
  \caption{Qualitative comparison of intermediate BEV feature representations for a VoD validation scene Id - 00000.}
  \label{fig:bev_qual}
\end{figure*}

\begin{figure*}[t]
  \centering
  \begin{minipage}[b]{0.26\linewidth}
    \centering
    \includegraphics[width=\linewidth, trim={0mm 0mm 0mm 0mm}, clip]{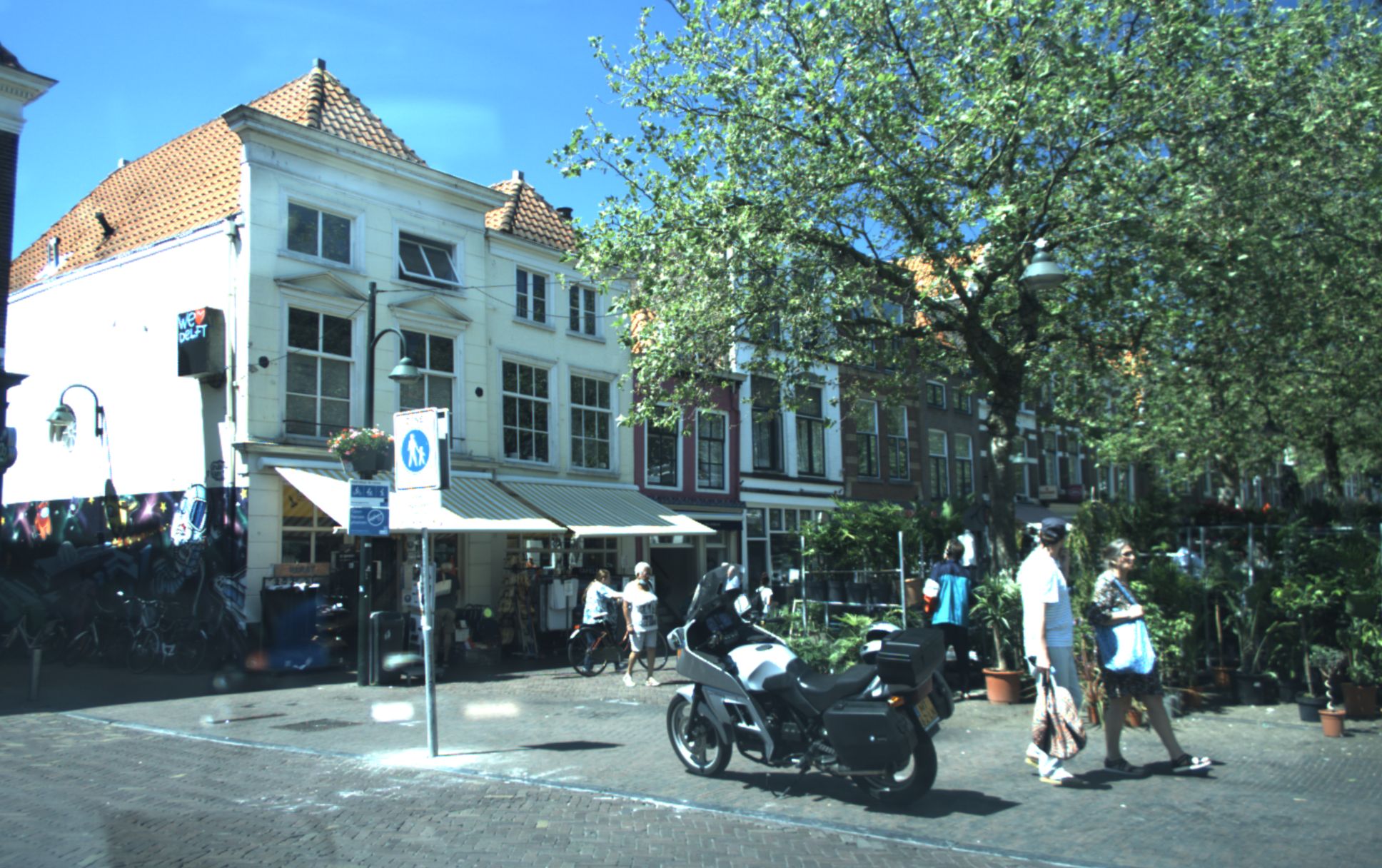}
    \vspace{2pt}
    \small (a) Validation Scene Id - 00033
  \end{minipage}
  \hfill
  \begin{minipage}[b]{0.24\linewidth}
    \centering
    \includegraphics[width=\linewidth, trim={0mm 75mm 35mm 60mm}, clip]{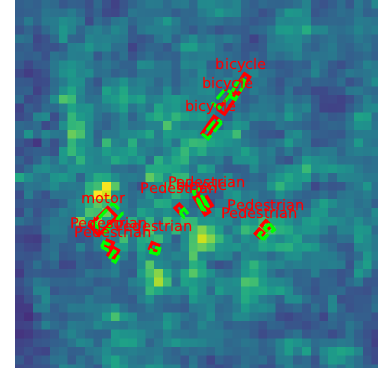}
    \label{fig:bev_qual_1_b}
    \vspace{2pt}
    \small (b) $F_c'$ - DCA Camera
  \end{minipage}
  \hfill
  \begin{minipage}[b]{0.24\linewidth}
    \centering
    \includegraphics[width=\linewidth, trim={0mm 75mm 35mm 60mm}, clip]{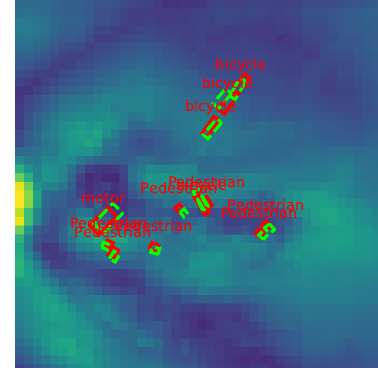}
    \label{fig:bev_qual_1_c}
    \vspace{2pt}
    \small (c) $F_r'$ - DCA Radar
  \end{minipage}
  \hfill
  \begin{minipage}[b]{0.24\linewidth}
    \centering
    \includegraphics[width=\linewidth, trim={0mm 75mm 35mm 60mm}, clip]{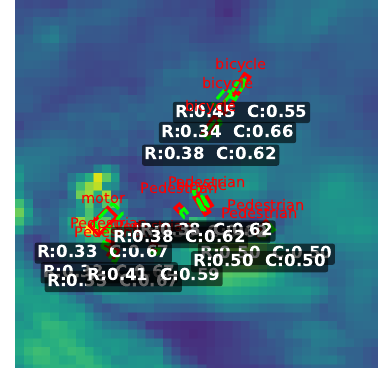}
    \label{fig:bev_qual_1_d}
    \vspace{2pt}
    \small (d) $F_f$ - Fused Features
  \end{minipage}
  \caption{Qualitative comparison of intermediate BEV feature representations for VoD validation scene Id - 00033.}
  \label{fig:bev_qual_1}
\end{figure*}

\begin{figure*}[t]
  \centering
  \begin{minipage}[b]{0.26\linewidth}
    \centering
    \includegraphics[width=\linewidth, trim={0mm 0mm 0mm 0mm}, clip]{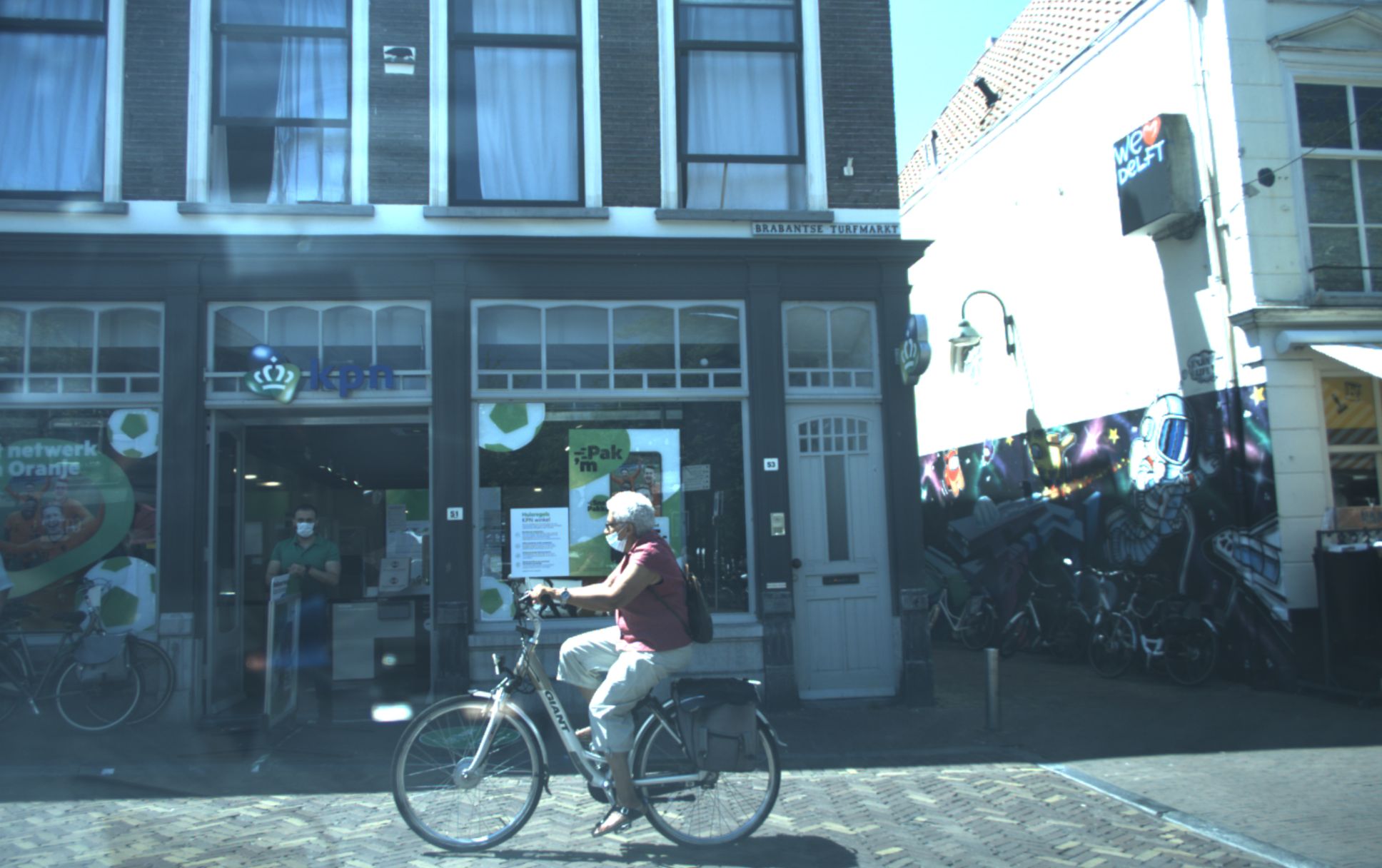}
    \vspace{2pt}
    \small (a) Validation Scene Id - 00102
  \end{minipage}
  \hfill
  \begin{minipage}[b]{0.24\linewidth}
    \centering
    \includegraphics[width=\linewidth, trim={0mm 75mm 35mm 60mm}, clip]{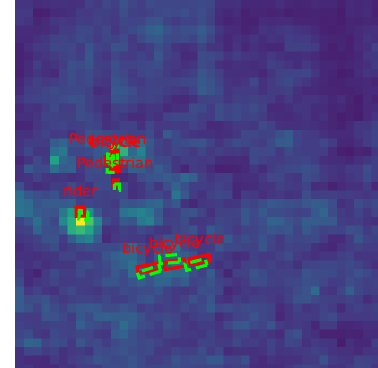}
    \label{fig:bev_qual_2_b}
    \vspace{2pt}
    \small (b) $F_c'$ - DCA Camera
  \end{minipage}
  \hfill
  \begin{minipage}[b]{0.24\linewidth}
    \centering
    \includegraphics[width=\linewidth, trim={0mm 75mm 35mm 60mm}, clip]{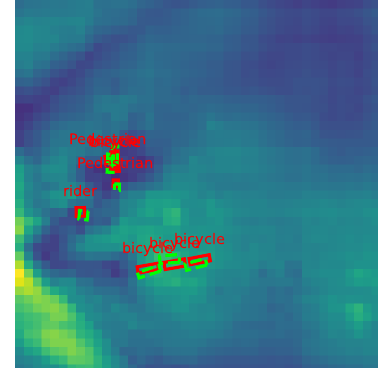}
    \label{fig:bev_qual_2_c}
    \vspace{2pt}
    \small (c) $F_r'$ - DCA Radar
  \end{minipage}
  \hfill
  \begin{minipage}[b]{0.24\linewidth}
    \centering
    \includegraphics[width=\linewidth, trim={0mm 75mm 35mm 60mm}, clip]{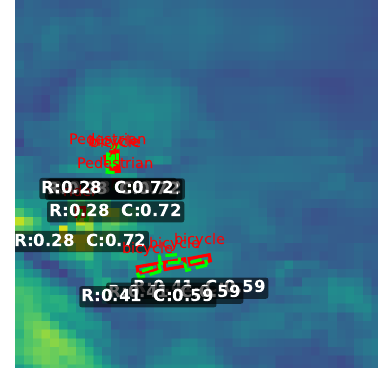}
    \label{fig:bev_qual_2_d}
    \vspace{2pt}
    \small (d) $F_f$ - Fused Features
  \end{minipage}
  \caption{Qualitative comparison of intermediate BEV feature representations for VoD validation scene Id - 00102.}
  \label{fig:bev_qual_2}
\end{figure*}

\begin{figure*}[t]
  \centering
  \includegraphics[width=0.8\linewidth, trim={0mm 0mm 0mm 6mm}, clip]{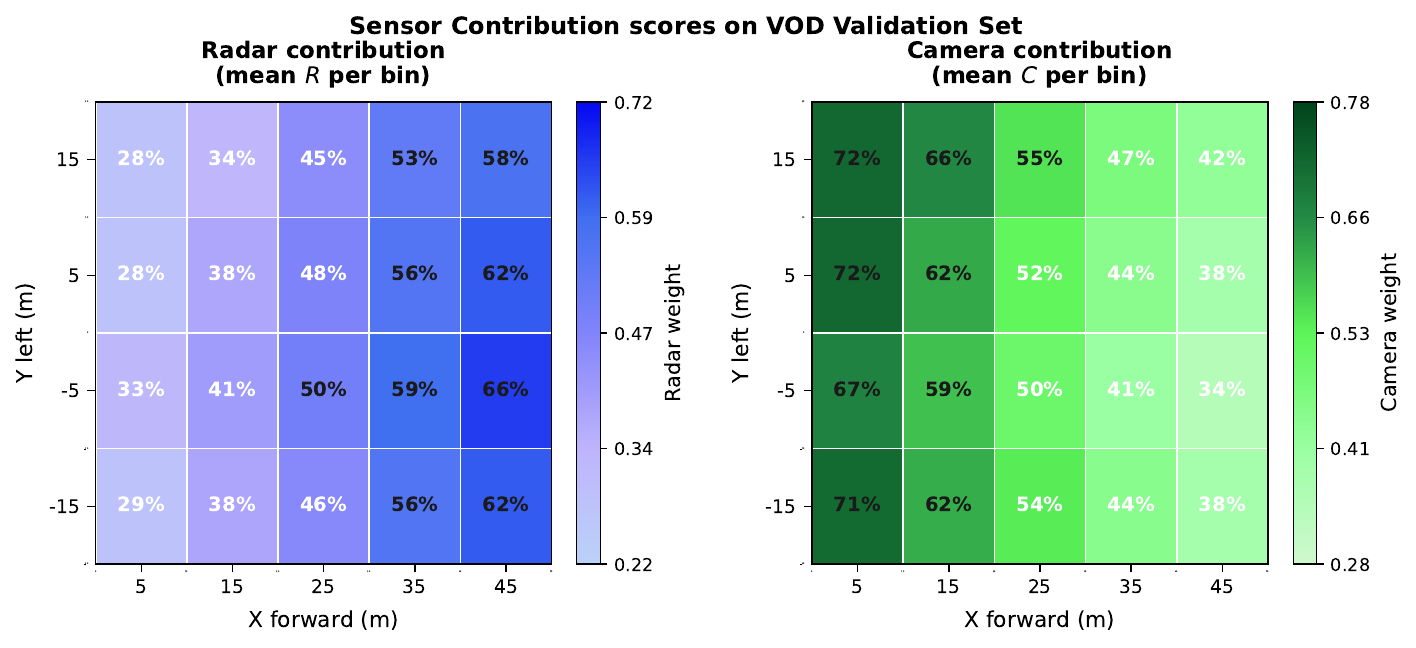}
  \caption{Sensor contribution maps on VoD validation set.}
  \label{fig:contrib_vod}
\end{figure*}

\textbf{Camera-only performance is limited by single-view depth estimation.}
The C-only configuration exhibits a substantial performance gap compared to radar-inclusive models, achieving 22.26\% mAP in the full area and 38.15\% in the ROI. This outcome is expected, as MMF-BEV relies on a single forward-facing monocular camera, in contrast to multi-camera setups commonly used in nuScenes-oriented pipelines. The limited perspective diversity restricts depth estimation accuracy, even with explicit depth supervision in BEVDepth. The particularly low pedestrian AP (15.78\% in the ROI) underscores the impact of monocular depth ambiguity, where objects at similar depths are projected into overlapping BEV bins, leading to fragmented detections. These findings highlight the importance of radar as a complementary geometric reference.

\textbf{Radar-only provides a strong spatial baseline.}
In contrast, the R-only branch attains 43.26\% mAP in the full area and
63.44\% in the ROI, substantially outperforming C-only and approaching
published fusion methods. This demonstrates the reliability of 4D radar measurements and the effectiveness of the RCS-aware BEV encoding strategy in stabilizing spatial localization. The Car AP of 67.21\% (ROI) is particularly strong,
consistent with cars producing high, stable RCS returns as used in RadarBEVNet.

\textbf{Naive concatenation reveals the cost of ignoring cross-modal 
alignment.}
MMF-BEV$^{*}$, which directly concatenates camera and radar BEV features 
without any attention mechanism, scores 45.45\% (full) and 65.62\% (ROI) 
— above camera-only but well below the full model. The pedestrian ROI AP 
drops sharply to 39.23\% versus 48.23\% for the full model, a relative 
degradation of 18.6\%. Without attention-guided alignment, the geometric 
mismatch between diffuse camera projections and sparse radar returns cannot 
be resolved. Notably, cyclist AP remains competitive (87.27\% vs. 87.20\%), 
suggesting that for larger, radar-reflective objects spatial alignment is 
less critical. This confirms that attention-based fusion is a functional 
requirement for small object detection, not merely an architectural choice.

\textbf{Fusion narrows the gap to published methods.}
MMF-BEV (C+R) achieves 48.92\% mAP over the full annotated area and 69.21\% within the ROI, yielding performance competitive with RCFusion (49.65\% / 69.23\%) and RCBEVDet (49.99\% / 69.80\%). In the ROI, MMF-BEV surpasses RCFusion in car AP and remains close to both prior fusion approaches overall, reducing the gap to RCBEVDet to less than one mAP point. MMF-BEV framework is inspired by the cross attention framework of RCBEVDet \cite{lin2024rcbevdet} and we have considered it as baseline to ensure the maintenance of the known performance. These results indicate that MMF-BEV achieves equivalent performance on VoD while introducing architectural modifications aimed at improving long-range fusion robustness and computational efficiency. Specifically, in-addition to the deformable cross attention, our design emphasizes structured deformable self-attention within each modality and enables each branch to function as a standalone detector, thereby increasing architectural flexibility.

The VoD dataset contains only a limited number of long-range annotated objects ($> 50\,\mathrm{m}$ in the longitudinal direction), which constrains the assessment of distant detection performance. To better evaluate long-range fusion capabilities and generalization, future work will train and benchmark the proposed architecture on datasets with richer long-range ground truth, such as the aiMotive \cite{matuszka2022aimotive} and NVIDIA Physical AI datasets \cite{nvlabs_physical_ai_av}. We are going to investigate further based on the architectural changes mentioned in our future work, through the help of additional internal and open-source datasets.

\subsection{Qualitative BEV Feature Map Visualization}

Figs.~\ref{fig:bev_qual}--\ref{fig:bev_qual_2} illustrate representative BEV feature maps extracted at three intermediate stages of MMF-BEV on VoD validation scenes. Ground-truth boxes are shown in green (dashed) and predictions in red. The contrast between the deformable cross-attention (DCA) branches highlights their complementary spatial characteristics. The image-query branch (Fig.~\ref{fig:bev_qual_b}) uses camera queries to attend to radar keys and values, resulting in a semantically rich but spatially diffuse BEV response; this is particularly noticeable around dense pedestrian clusters, where monocular depth ambiguity spreads activations across multiple depth bins. Conversely, the radar-query branch (Fig.~\ref{fig:bev_qual_c}) uses radar queries anchored at precise physical returns to sample image features, producing compact, well-localized activations concentrated around object centers. Objects with strong radar returns, such as trucks and cyclists, generate the most focused responses. The fused BEV map (Fig.~\ref{fig:bev_qual_d}) combines these effects, producing sharp, high-magnitude responses for large objects while pedestrian clusters remain comparatively diffuse, consistent with the pedestrian AP gaps reported in Table~\ref{tab:vod_roi}. 

\subsection{Sensor Contribution Analysis}
\label{subsec:contrib_results}

Fig.~\ref{fig:contrib_vod} presents sensor contribution maps computed on the VoD validation set. Each BEV bin visualizes the mean radar weight $R(i,j)$ (left) and the corresponding camera weight $C(i,j)=1-R(i,j)$ (right), averaged over all detections with the origin at LiDAR mounting position.

\begin{itemize}

    \item \textbf{Distance-stratified analysis.}
    Camera contributions dominate in the near-field region ($X < 15\,\mathrm{m}$), with $C \approx 65$--$72$\%, where objects occupy larger image regions and BEVDepth’s explicit depth supervision provides accurate projections. At greater distances ($X > 30\,\mathrm{m}$), radar contributions increase ($R > 55$\%), reflecting the growing uncertainty of monocular depth estimation and the higher reliability of radar point clouds for spatial localization specially in radial direction to the sensor.

    \item \textbf{Class-stratified analysis.}
    Radar contributions are higher for cars and cyclists, whose larger RCS returns and Doppler signatures provide reliable spatial anchors. Pedestrians show the strongest camera reliance, as their small RCS and irregular motion produce weak radar returns, making visual texture the more discriminative cue. 

\end{itemize}

These trends demonstrate that MMF-BEV’s cross-attention mechanism adaptively modulates modality weighting across BEV space, effectively leveraging the complementary characteristics of radar and camera features.

\section{Conclusion}
\label{sec:conclusion}

We presented MMF-BEV, a multi modal BEV fusion framework that systematically investigates deformable attention both within individual modalities and across modalities for 3D object detection on the View-of-Delft (VoD) 4D radar dataset. By equipping the camera and radar branches with Deformable Self-Attention and coupling them through a deformable cross-attention module in BEV space, the proposed architecture enables structured and spatially adaptive information exchange between modalities.

Experimental results on the VoD validation set demonstrate that cross-modal fusion consistently outperforms both unimodal variants across the entire annotated area and the Region of Interest. The substantial gains over the camera-only and radar-only branches confirm that the two sensing modalities provide complementary geometric and semantic cues. The sensor contribution analysis reveals a spatially structured, interpretable pattern of modality-specific contributions in BEV space: camera features dominate at short ranges, where monocular depth estimation is most reliable, while radar features increasingly contribute at longer distances, particularly for dynamic objects and off-axis locations where camera depth uncertainty accumulates. This distance- and location-dependent complementarity directly validates the design rationale of the deformable cross-attention mechanism, which explicitly queries each modality from the perspective of the other.

Future work will investigate alternative radar backbones tailored to 4D radar representations, improved image-based depth estimation strategies for the camera branch, and more advanced cross-modal transformer designs, including adaptive and uncertainty-aware attention mechanisms. In addition, we plan to extend the framework to multi-view camera setups on other datasets and benchmark its performance to further assess generalization and scalability.

\bibliographystyle{IEEEtran}
\bibliography{references}

\end{document}